\documentclass[11pt]{article}

\usepackage[preprint]{acl}

\usepackage{times}
\usepackage{latexsym}

\usepackage[T1]{fontenc}
\usepackage[utf8]{inputenc}

\usepackage{microtype}
\usepackage{inconsolata}
\usepackage{graphicx}
\usepackage{booktabs}
\usepackage{listings}

\title{Verdicts Without Annotated Evidence: Rejection Sampling or Label-Only Post-Training for Evidence Recovery?}

\author{
  Nishanth Nayakanti,
  Prasang Gupta,
  Ashutosh Bilthare,
  Kevin Paul
\\
  PricewaterhouseCoopers, U.S.
\\
  \small{
    \textbf{Correspondence:} \href{mailto:nishanth.nayakanti@pwc.com}{nishanth.nayakanti@pwc.com}
  }
}

\begin{document}
\maketitle

\begin{abstract}
In many review workflows the verdict is the only thing retained. The passages behind it are not marked, because that annotation costs far more than recording the decision. We measure how much of that evidence a small language model can recover when it is post-trained on the verdicts alone, with no human evidence labels at any stage. On ContractNLI the human evidence spans are held out until evaluation. Matching the recorded verdict and agreeing with those spans are not the same thing: across six systems the two scores are only weakly related and rank the systems differently, so accuracy is a poor guide when the citations have to be reviewable. Label-only training on the bare verdict reaches accuracy 0.896 and span F1 0.564. Rejection sampling, which keeps a generated trace only when its verdict matches the record and then picks one by an automatic source-grounding score, reaches 0.797 and 0.556, against 0.747 and 0.493 before training. Verbatim citation rises from 0.597 to 0.729 under label-only training and to 0.701 under rejection sampling. One seed on one corpus cannot say which method is better, but both improve the evidence without anyone \mbox{annotating it}.

\end{abstract}

\section{Introduction}
\label{sec:intro}

A professional reviewing a document against a standard produces a verdict. In many operational workflows that verdict is the only artifact retained: the reasoning is not recorded and the supporting passages are not marked. This asymmetry follows from how the work is recorded, not from any particular domain.

Supervised systems for these tasks are commonly built from richer records --- for each document--question pair, annotators mark the passages that support the recorded conclusion. Producing those span-level labels is far more expensive than retaining the verdict alone, and operational records rarely contain them. The question is narrow: if only verdicts are available for training, how much evidence quality can be recovered without evidence annotations, and what actually does the work?

\textbf{Two axes, not one.} We separate \emph{verdict accuracy} (did the model reach the recorded conclusion?) from \emph{evidence quality} (are the passages it cites verifiable and aligned with withheld human evidence?). These are distinct properties, and Section~\ref{sec:decouple} shows they rank systems differently --- weakly enough that the most accurate off-the-shelf system we evaluate (gpt-5.2, 0.808) is no better on evidence than the untrained student, span F1 0.493 for both, while two lower-accuracy systems --- the 31B sampler and sonnet-4.6 --- score highest on evidence. A workflow that needs reviewable evidence cannot select a model on accuracy alone.

\textbf{The shape, not the domain.} ContractNLI \citep{koreeda2021contractnli} is useful here not because it concerns contracts but because it reproduces the operational structure: a fixed set of questions applied to every document, one recorded verdict per question, and a separate evidence layer that can be withheld from training. The first two properties are common in operational records; the third is usually absent. Checklists against filings, claims against policies, submissions against standards, and cases coded against schedules all retain conclusions while omitting the passages behind them. ContractNLI makes the question measurable precisely because its evidence layer can be hidden during training and used only to evaluate what was learned without it. It is a proxy, not the operational setting itself: what we report is one corpus, one student, one seed per arm.

\textbf{Two methods, one question.} We compare two post-training methods that both use only the recorded verdict --- no human evidence annotation anywhere:

\begin{itemize}
\item \textbf{Label-only SFT:} fine-tune the student directly on the bare recorded verdicts --- the direct-answer baseline these methods are usually measured against \citep{hsieh2023distilling}.
\item \textbf{Rejection-sampling SFT (RSFT):} following STaR \citep{NEURIPS2022_639a9a17}, generate candidate traces --- from the student and a larger 31B sampler --- keep only those whose verdict matches the record, select one per row by an automatic source-grounding score, and fine-tune on the selected rationale-plus-citation traces. We write RSFT for our instance throughout: rejection-sampling fine-tuning (RFT; \citealp{yuan2023scalingrelationshiplearningmathematical}) with an added source-grounding selection stage over the surviving candidates, which RFT does not have. (About a fifth of the selected traces come from the sampler; Section~\ref{sec:rsft}, Table~\ref{tab:sources}.)
\end{itemize}

RSFT is the more elaborate method --- it costs an extra candidate-generation stage and a grounding selector, and it trains the model to \emph{emit} grounded citations. Label-only SFT is the cheapest thing that could work: it skips candidate generation and selection entirely, so end-to-end it is a single \textasciitilde{}2.6 GPU-hour training run against RSFT's \textasciitilde{}9.7 (6.8 for candidate generation plus 2.9 for training; Section~\ref{sec:rsft}, Appendix~\ref{sec:generation}). Comparing them provides a practical test of what additional value rationale-plus-citation supervision provides over bare-label supervision.

\textbf{What we find.} Both methods recover evidence quality without evidence annotation, and we report them as a comparison rather than a ranking. In this single-seed, single-corpus setting the two arms land close on the evidence axis (span F1 0.556 for RSFT, 0.564 for label-only, against the untrained student's 0.493) and further apart on accuracy (0.797 vs 0.896, against 0.747). Both also raise source verifiability --- the rate at which a cited passage appears verbatim in the contract --- from 59.7\% untrained to 70.1\% for RSFT and 72.9\% for label-only. A 2-epoch RSFT run reproduces the 1-epoch result (0.796), arguing against insufficient epoch count as the explanation for the gap. A candidate mechanism ties the two axes together: evidence F1 is mediated by the verdict --- a NotMentioned prediction returns no spans, and a correct Entailment/Contradiction prediction returns citations the base model already knows how to produce --- so verdict-improving fine-tuning can surface more evidence. Read together, the results are consistent with the base model already containing substantial evidence-extraction capability that verdict-only adaptation surfaces, rather than capability the supervision installs. On this reading the evidence gain does not appear to require rationale supervision, but a single seed cannot settle which method is preferable, and we leave that to the stacked and repeated-seed variants in Section~\ref{sec:roadmap}.

Our contributions are as follows:
\begin{enumerate}
\item We frame evidence recovery from verdict-only records as a measurable post-training problem and separate two quality axes --- verdict accuracy and evidence-span agreement --- that we show decouple across systems (Section~\ref{sec:decouple}).
\item We compare two verdict-only post-training methods under a matched decode and a locked prompt, with one greedy, unfiltered output per model at evaluation and no human evidence annotation at any stage (Sections~\ref{sec:methods}--\ref{sec:evidence}).
\item We show the two methods reach comparable evidence quality by different routes: label-only SFT --- trained on bare labels, with no rationale or citation --- gains through the verdict, while RSFT gains through the citation quality its source-grounded supervision buys, its verdicts actually dipping on the scored rows (Section~\ref{sec:evidence}). Evidence gains here need no evidence annotation, and for label-only no rationale supervision at all; we do not claim one method is preferable in general.
\item We bound alternative explanations: a label-marginal calibration control --- fit on the training split, reading no contract text and requiring no training of its own --- recovers only 37\% of RSFT's accuracy increment (Section~\ref{sec:calibration}), and a 2-epoch RSFT run reproduces the 1-epoch result, arguing against insufficient epoch count (Section~\ref{sec:config}).
\end{enumerate}

\section{Setup}
\label{sec:setup}

\subsection{Task and dataset}
\label{sec:task}

\textbf{Task.} ContractNLI \citep{koreeda2021contractnli} pairs each contract with 17 fixed hypotheses. For each document--hypothesis pair the system predicts one of three labels: Entailment when the contract supports the hypothesis, Contradiction when it conflicts, and NotMentioned when the contract provides neither. Human-marked evidence spans accompany entailment and contradiction rows; they are withheld from prompting, candidate filtering and selection, training, and checkpoint choice, and are used only for evaluation.

\begin{figure*}[t]
  \centering
  \includegraphics[width=\linewidth]{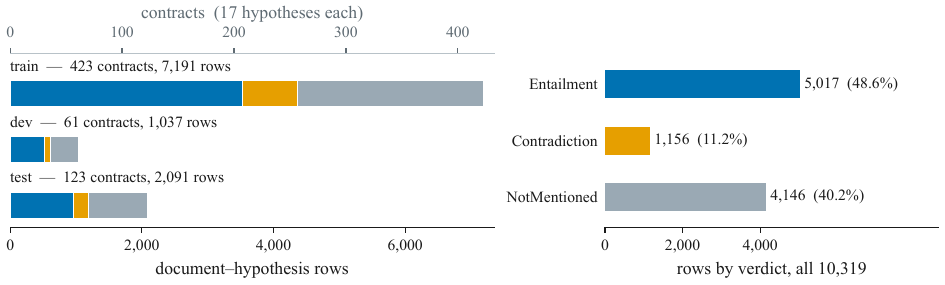}
  \caption{The corpus. Every contract is assessed against all 17 hypotheses, so rows are document--hypothesis pairs. The verdict distribution is over all 10,319 rows; the test split's NotMentioned rate is higher, at 43.2\%. Evaluation is on the canonical test split: 123 contracts, 2,091 rows.}
  \label{fig:corpus}
\end{figure*}

\textbf{Withheld annotation layer.} Each document is pre-segmented into candidate spans (median 71, mean 79) and each row marks every span as evidence or not. On the training split this yields 559,215 binary span decisions alongside 7,191 verdicts. We use the verdicts to construct supervision and reserve the span labels for held-out measurement, matching the operational asymmetry between routinely retained decisions and separately commissioned evidence annotation.

\subsection{Models and controls}
\label{sec:models}

\begin{table*}[t]
  \centering
  \small
  \begin{tabular}{@{}p{0.14\textwidth}p{0.24\textwidth}p{0.56\textwidth}@{}}
    \toprule
    \textbf{Role} & \textbf{System} & \textbf{Uses in this study} \\
    \midrule
    Student & Qwen3.5-9B \citep{qwen35}, 4-bit, QLoRA & Post-trained by both methods; evaluated before and after \\
    Sampler & Gemma-4-31B \citep{gemma4} & Supplies additional candidate traces to RSFT; also an inference reference \\
    Commercial references & sonnet-4.6 \citep{anthropic-sonnet46}, gpt-5.2 \citep{openai-gpt52} & Inference-time comparisons only; no training data or supervision \\
    Method A & Label-only SFT & Fine-tune the student on bare recorded verdicts \\
    Method B & RSFT & Fine-tune on verdict-filtered, source-grounded traces from the student and a 31B sampler \\
    Control & Prior calibration & Relabel the untrained student's predictions from training-split confusions (Section~\ref{sec:calibration}) \\
    Control & Hypothesis lookup & Predict from hypothesis identity, reading no contract (0.659 accuracy) \\
    Control & Majority class & 0.463 accuracy \\
    \bottomrule
  \end{tabular}
  \caption{Models, methods, and controls.}
  \label{tab:models}
\end{table*}

Two non-neural controls contextualize verdict accuracy. The majority-class predictor scores 0.463. The hypothesis-only lookup control predicts from hypothesis identity without reading the contract and scores 0.659; because the task repeats 17 hypotheses, it estimates how much performance comes from hypothesis-specific label priors --- the same class-prior artifact that hypothesis-only baselines expose in NLI datasets \citep{poliak2018hypothesis,gururangan2018artifacts}. Performance above it is consistent with using information beyond those priors, though it does not by itself identify how the contract text is used.

\textbf{Commercial reference identifiers.} \texttt{sonnet-4.6} is Anthropic Claude Sonnet 4.6 on AWS Bedrock (\texttt{us.anthropic.claude-sonnet-4-6}, \texttt{us-east-1}, undated \texttt{v1} snapshot); \texttt{gpt-5.2} is OpenAI \texttt{gpt-5.2-2025-12-11}. Both were queried under greedy decoding on 23 September 2026; neither records completion-token counts (Section~\ref{sec:eval}), so their decode caps cannot be recovered from the saved records.

\subsection{Prompt and output format}
\label{sec:prompt}

\textbf{Two prompts, one task.} A single task definition is locked and hash-checked across systems and runs. We use two output variants of it:

\begin{itemize}
\item \textbf{\texttt{faithful}} (hash \texttt{f9318d34}) --- asks for the verdict plus concise verbatim evidence quotations and a short reasoning field. This is the prompt used for every evidence-axis evaluation and by RSFT candidate generation.
\item \textbf{\texttt{verdict\_only}} (hash \texttt{7ada9e8d}) --- the same contract, hypothesis, and three-way task instruction, with the evidence paragraph removed and the output reduced to the bare verdict. Used to train and evaluate label-only SFT on the accuracy axis.
\end{itemize}

Everything above the output block is byte-identical between the two except the single evidence paragraph; only the requested output differs (full text in Appendix~\ref{sec:appendix}). We drop the evidence paragraph from \texttt{verdict\_only} because leaving it in would instruct a behaviour the label-only arm has no target for --- training the model to disobey its own prompt --- the confound STaR \citep{NEURIPS2022_639a9a17} and Distilling Step-by-Step \citep{hsieh2023distilling} avoid by adapting the direct-answer prompt. This choice creates a deliberate train/test mismatch for the label-only arm: it is trained under \texttt{verdict\_only} and evaluated on the evidence axis under \texttt{faithful}, a prompt it never saw in training. The mismatch runs against that arm rather than for it, and its cost on the accuracy axis is small --- 0.896 under \texttt{faithful} against 0.899 under \texttt{verdict\_only} (Table~\ref{tab:verdict}). That a model trained to emit bare verdicts nonetheless cites on 93.4\% of rows when asked is part of the Section~\ref{sec:evidence} result, not an artifact of it.

The template adds no worked examples, retrieval stage, or decision rules. A tested variant containing three plausible decision rules changed none of the polarity errors it was written to fix (a disclosed single observation, not a controlled study). Structured output is the one imposed requirement: the citation must arrive as a field, or the evidence measure is not computable. The \texttt{faithful} prompt also elicits a 1--10 confidence score; it is near-degenerate in practice, we do not analyse it, and it enters no reported metric.

\subsection{Evaluation protocol}
\label{sec:eval}

All model evaluations use the canonical test split of 123 contracts and 2,091 rows. Each system produces one greedy, single-pass output per row under the shared prompt; no inference-time candidate filtering or selection is applied. Unparseable outputs are scored as incorrect; outputs without a citation receive no predicted evidence spans. Verdict accuracy and macro F1 are reported on all rows. Evidence precision, recall, and F1 are computed on the fixed 1,188 test rows whose gold label is Entailment or Contradiction --- the classes for which ContractNLI provides human evidence spans --- regardless of the model's predicted label. A model that predicts NotMentioned and emits no evidence on an eligible row incurs no false-positive spans but misses the gold spans, reducing recall and F1.

\textbf{Matched decode.} All reported runs use a 900-token completion cap with one exception. An earlier release decoded the untrained student and the 31B sampler at 700 tokens while the post-trained student ran at 900; we re-ran the untrained student at 900 (Section~\ref{sec:config}), so every comparison involving the two fine-tunes and their baseline uses an identical cap. The 31B sampler was not re-run and its figures are from the 700-token pass; only 1 of its 2,091 rows reached that cap, so they are effectively cap-insensitive (Table~\ref{tab:decoupling}). The two commercial systems do not record completion-token counts, so their caps cannot be read from the saved records.

\section{Methods compared}
\label{sec:methods}

Both methods post-train the same student with low-rank adapters \citep{hu2022lora} over a 4-bit quantised base (QLoRA; \citealp{NEURIPS2023_1feb8787}) for one epoch, loss on target tokens only, no human evidence annotation, and no checkpoint selection. They differ in the supervision target and the corresponding training-output instruction --- label-only is trained and prompted with \texttt{verdict\_only}, RSFT with \texttt{faithful} --- and evidence quality is compared under the matched \texttt{faithful} prompt (Section~\ref{sec:evidence}). RSFT is an instance of rejection-sampling / reward-ranked fine-tuning \citep{yuan2023scalingrelationshiplearningmathematical,dong2023raft}; label-only SFT is the direct-answer baseline against which rationale-supervised methods are usually measured \citep{hsieh2023distilling}. Section~\ref{sec:related} places both in the wider literature.

\subsection{Method A --- Label-only SFT}
\label{sec:label-only}

The target is the bare recorded verdict, wrapped in the \texttt{<answer>} block the parser expects and nothing else:

\begin{lstlisting}
<answer>
{"hypothesis_id": "nda-2", "answer": "Contradiction"}
</answer>
\end{lstlisting}

Targets are reduced to a mean of 71 characters (down from 1,294 for the rationale traces). Training uses 7,191 rows on a 6,477 / 714 document-level split; the training documents share zero documents and zero (document, hypothesis) rows with the test split. No evidence annotation, no rationale, no candidate generation.

\subsection{Method B --- Rejection-sampling SFT (RSFT)}
\label{sec:rsft}

RSFT generates multiple candidate traces per training row, rejects those whose verdict disagrees with the record, ranks survivors by an automatic source-grounding score, selects one target per covered row, and fine-tunes on the assembled set. Verdict filtering determines eligibility; grounding selection chooses among eligible candidates. No stage reads human evidence annotations.

\begin{figure*}[t]
  \centering
  \includegraphics[width=\linewidth]{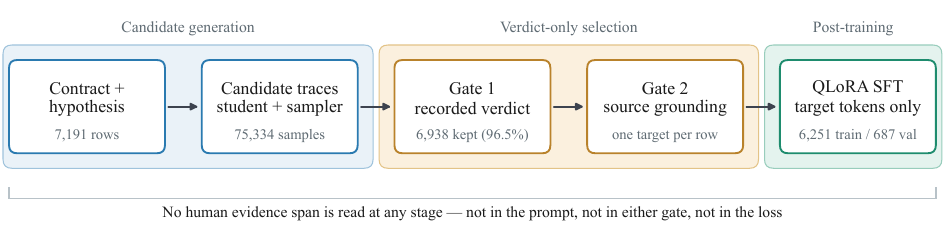}
  \caption{RSFT pipeline: candidate generation $\rightarrow$ verdict filtering $\rightarrow$ source-grounding selection $\rightarrow$ one epoch of target-only QLoRA.}
  \label{fig:method}
\end{figure*}

\textbf{Candidate generation (Appendix~\ref{sec:generation}).} Both models produce one greedy candidate per training row; sampled passes then run only on rows whose greedy output fails the verdict filter --- 32 samples for the student, 8 for the sampler. Candidates are pooled by row before filtering. Candidate generation costs 6.8 GPU-hours; training costs 2.9.

\begin{table}[t]
  \centering
  \footnotesize
  \setlength{\tabcolsep}{4pt}
  \begin{tabular}{@{}p{0.28\columnwidth}p{0.15\columnwidth}p{0.16\columnwidth}p{0.23\columnwidth}@{}}
    \toprule
    \textbf{Pass} & \textbf{Rows attempted} & \textbf{Samples} & \textbf{Rows with $\geq$1 accepted} \\
    \midrule
    Greedy, Gemma-4-31B & 7,191 & 7,191 & 5,908 (82.2\%) \\
    Greedy, Qwen3.5-9B & 7,191 & 7,191 & 5,607 (78.0\%) \\
    Sampled, Gemma-4-31B (k=8) & 1,283 & 10,264 & 245 (19.1\%) \\
    Sampled, Qwen3.5-9B (k=32) & 1,584 & 50,688 & 1,248 (78.8\%) \\
    Pooled & 7,191 & 75,334 & 6,938 (96.5\%) \\
    \bottomrule
  \end{tabular}
  \caption{\textbf{Gate 1 (verdict filter) budget and yield.} The sampled passes run only on rows their own greedy pass failed, so their denominators are residues, not the corpus.}
  \label{tab:gate1}
\end{table}

As shown in Table~\ref{tab:gate1}, the two greedy passes cover 86.2\% of rows; the sampled passes add 741, reaching 96.5\%; the remaining 253 rows are dropped.

\textbf{Verdict filtering.} Every candidate is retained only when its predicted verdict equals the recorded label --- no partial credit, no confidence threshold, no inspection of the rationale.

\textbf{Source-grounding selection.} Survivors are ranked from their quoted excerpts alone. \emph{Step 1, well-formedness:} a candidate is eligible if it closes an \texttt{<answer>} block, parses as JSON, has a non-empty reasoning field, and follows the citation rule (evidence for E/C, none for NotMentioned); this removes 2,746 candidates and leaves $\geq$1 well-formed candidate on 98.6\% of covered rows. On the remaining 1.4\% no verdict-matching candidate is well-formed; rather than drop the row, the method keeps its raw verdict-matching pool and scores that, so every covered row still yields a target and Table~\ref{tab:sources} totals all 6,938. \emph{Step 2, grounding:} each excerpt scores 3 (verbatim), 2 (verbatim after normalisation), 1 ($\geq$60\% of content tokens in one contiguous stretch), or 0 (unlocatable); the candidate score is the mean excerpt score over 3. The highest-scoring candidate wins; ties use a fixed provenance order (student greedy, student sampled, sampler greedy, sampler sampled).

\textbf{Note on the grounding instrument.} Selection and evaluation share the same
three-tier span-matching ladder (Section~\ref{sec:evidence}), so span F1 is partly a
locatability measure; we therefore lead with verbatim rate in
Section~\ref{sec:evidence}. Table~\ref{tab:selection} quantifies what the selector is
worth in the training archive. The provenance tie-break costs 0.013 span F1 relative
to load-order ties, and raises the on-policy share from 21.5\% to 81.6\% in this re-selection (80.2\% in the shipped set, Table~\ref{tab:sources}). Nothing in either gate reads a human evidence label, and
nothing selects at inference.

\begin{table}[t]
  \centering
  \footnotesize
  \setlength{\tabcolsep}{4pt}
  \begin{tabular}{@{}p{0.40\columnwidth}rrrr@{}}
    \toprule
    \textbf{Selection rule} & \textbf{P} & \textbf{R} & \textbf{F1} & \textbf{Any hit} \\
    \midrule
    First survivor, no grounding & 0.915 & 0.534 & 0.675 & 83.0\% \\
    Grounding, tie by load order & 0.910 & 0.578 & 0.707 & 90.4\% \\
    Grounding, tie by provenance (shipped) & 0.890 & 0.569 & 0.694 & 90.0\% \\
    \midrule
    Oracle (reads gold spans) & 0.936 & 0.593 & 0.726 & 91.1\% \\
    \bottomrule
  \end{tabular}
  \caption{\textbf{What selection is worth in the training data.} Agreement of
  selected RSFT training targets with withheld human spans over 4,303 covered
  training rows. ``Any hit'' is the share of rows whose selected trace cites at
  least one correct span. The oracle picks, from the same candidate pool, the
  candidate with the highest agreement (headroom only). Gold spans are used only
  for this analysis: no stage reads them, and nothing selects at inference.}
  \label{tab:selection}
\end{table}

\begin{table*}[t]
  \centering
  \small
  \begin{tabular}{@{}lrrrr@{}}
    \toprule
    \textbf{How the row was decided} & \textbf{Rows} & \textbf{Share} & \textbf{Student wins} & \textbf{Sampler wins} \\
    \midrule
    One candidate left, kept as-is & 529 & 7.6\% & 347 (65.6\%) & 182 (34.4\%) \\
    Grounding score separated them & 1,513 & 21.8\% & 360 (23.8\%) & 1,153 (76.2\%) \\
    Grounding tied, provenance decided & 4,896 & 70.6\% & 4,858 (99.2\%) & 38 (0.8\%) \\
    Total & 6,938 & & 5,565 (80.2\%) & 1,373 (19.8\%) \\
    \bottomrule
  \end{tabular}
  \caption{Gate 2 (grounding) decision accounting.}
  \label{tab:gate2}
\end{table*}

Grounding separates candidates on 21.8\% of covered rows; on 70.6\% the survivors tie (usually because all quotations match verbatim) and the provenance order decides. Where grounding discriminates, the 31B sampler supplies 76.2\% of selected traces.

\begin{table}[t]
  \centering
  \footnotesize
  \setlength{\tabcolsep}{4pt}
  \begin{tabular}{@{}p{0.48\columnwidth}p{0.14\columnwidth}p{0.16\columnwidth}@{}}
    \toprule
    \textbf{Source of the selected trace} & \textbf{Rows} & \textbf{Share} \\
    \midrule
    Qwen3.5-9B, greedy & 4,382 & 63.2\% \\
    Gemma-4-31B, greedy & 1,317 & 19.0\% \\
    Qwen3.5-9B, sampled & 1,183 & 17.1\% \\
    Gemma-4-31B, sampled & 56 & 0.8\% \\
    Student total (on-policy) & 5,565 & 80.2\% \\
    \bottomrule
  \end{tabular}
  \caption{Where the 6,938 RSFT training targets came from.}
  \label{tab:sources}
\end{table}

Four fifths of targets are the student's own generations. Of the 5,025 excerpts in selected traces, 89.9\% match verbatim, 1.9\% after normalisation, 8.2\% partially, and 0.1\% cannot be located; mean excerpts per target is 0.72 because NotMentioned targets correctly contain no citation. These statistics describe the supervision, not a causal attribution.

\subsection{Controls}
\label{sec:controls}

\textbf{Prior calibration (Section~\ref{sec:calibration})} relabels the untrained student's own predictions through a confusion table fit on the training split, correcting the label marginal without reading a contract or training anything. Hypothesis lookup and majority class bound what hypothesis identity and the corpus rate alone can do.

\section{Verdict prediction results}
\label{sec:verdict}

\subsection{Aggregate performance}
\label{sec:aggregate}

\begin{table*}[t]
  \centering
  \small
  \begin{tabular}{@{}lllrr@{}}
    \toprule
    \textbf{System} & \textbf{Method} & \textbf{Prompt} & \textbf{Accuracy} & \textbf{Macro F1} \\
    \midrule
    Majority class & --- & --- & 0.463 & 0.211 \\
    Lookup control (reads no contract) & --- & --- & 0.659 & 0.617 \\
    Qwen3.5-9B, untrained & --- & faithful & 0.747 & 0.701 \\
    Qwen3.5-9B, untrained & --- & verdict\_only & 0.753 & 0.718 \\
    sonnet-4.6 & --- & faithful & 0.771 & 0.718 \\
    Gemma-4-31B (sampler) & --- & faithful & 0.786 & 0.748 \\
    Qwen3.5-9B, RSFT & RSFT & faithful & 0.797 & 0.758 \\
    gpt-5.2 & --- & faithful & 0.808 & 0.771 \\
    Qwen3.5-9B, label-only & label-only & faithful & 0.896 & 0.860 \\
    Qwen3.5-9B, label-only & label-only & verdict\_only & 0.899 & 0.861 \\
    \bottomrule
  \end{tabular}
  \caption{Held-out verdict prediction, 2,091 test rows, matched 900-token decode. Label-only SFT is reported under both prompts; RSFT and the untrained student under \texttt{faithful}.}
  \label{tab:verdict}
\end{table*}

\textbf{The two fine-tunes rank in the opposite order to their cost.} Label-only SFT reaches 0.896--0.899, above RSFT's 0.797 by roughly +0.10 and above its prompt-matched base (0.753) by +0.145, at one epoch each. It sits well above the hypothesis-identity lookup control (0.659), consistent with using information beyond hypothesis-specific priors, though this does not by itself identify the mechanism or exclude pretraining exposure (Section~\ref{sec:secondary}, \hyperref[sec:limitations]{Limitations}). The shift is not majority-gaming: NotMentioned recall moves 0.59 $\rightarrow$ 0.895, all three class recalls rise, and the predicted mix lands on the gold marginal (predicted NM 0.425 vs gold 0.432).

\begin{table*}[t]
  \centering
  \small
  \setlength{\tabcolsep}{4pt}
  \begin{tabular}{@{}p{0.20\textwidth}lllp{0.23\textwidth}@{}}
    \toprule
    \textbf{Comparison} & \textbf{$\Delta$ accuracy} & \textbf{95\% CI (row)} & \textbf{95\% CI (doc-clustered)} & \textbf{McNemar} \\
    \midrule
    RSFT vs untrained (matched decode) & +0.050 & [+0.036, +0.064] & [+0.035, +0.065] & 166 fixed, 62 broken, p = $4\times 10^{-12}$ \\
    Label-only vs untrained & +0.149 & [+0.131, +0.167] & [+0.129, +0.169] & 359 fixed, 48 broken, p = $6\times 10^{-60}$ \\
    Label-only vs RSFT & +0.099 & [+0.083, +0.115] & [+0.083, +0.115] & 259 fixed, 52 broken, p = $3\times 10^{-34}$ \\
    RSFT vs 31B sampler & +0.011 & [$-$0.005, +0.027] & --- & p = 0.20 \\
    RSFT vs gpt-5.2 & $-$0.011 & [$-$0.027, +0.006] & --- & p = 0.21 \\
    \bottomrule
  \end{tabular}
  \caption{Paired accuracy comparisons (10,000 paired bootstrap resamples \citep{tibshirani1993introduction}, seed 37; exact two-sided McNemar \citep{McNemar_1947} on discordant pairs). CIs are reported both row-level and document-clustered (resampling the 123 test contracts, carrying all rows of each).}
  \label{tab:paired}
\end{table*}

\textbf{Clustering does not widen the intervals.} The document-clustered bootstrap \citep{tibshirani1993introduction} is nearly identical to the row-level one because the measured one-way ICC of the correctness indicator across documents is $\approx$ 0 (design effect $\approx$ 1.0 for all three fine-tunes) --- the row-level intervals the study reports are not anti-conservative here, which addresses the clustering concern directly rather than by assumption. The RSFT student is not statistically distinguishable from the 31B sampler or gpt-5.2 on accuracy; both intervals include zero. A non-significant difference is not evidence of equivalence --- we run no equivalence test (TOST; \citealp{Schuirmann1987,doi:10.1177/1948550617697177}) and do not describe the RSFT student as ``level with'' the references. The label-only-vs-RSFT accuracy gap is large relative to these intervals but remains a single-seed estimate (Section~\ref{sec:config}, \hyperref[sec:limitations]{Limitations}).

\begin{figure}[t]
  \centering
  \includegraphics[width=\columnwidth]{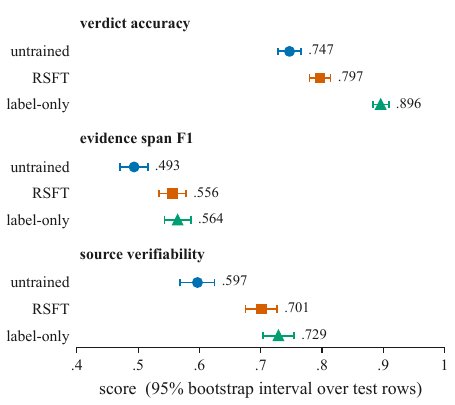}
  \caption{The two verdict-only methods against the untrained student, matched decode, \texttt{faithful} prompt; source verifiability is the all-verbatim rate of Section~\ref{sec:evidence}. Error bars are 95\% bootstrap intervals over each system's own test rows --- marginal, not paired --- and the horizontal axis is truncated. The two fine-tunes separate on verdict accuracy alone: their span-F1 intervals overlap, and so do their verifiability intervals (RSFT [0.675, 0.726] vs label-only [0.704, 0.754]). Overlapping marginal intervals do not settle a comparison; the paired difference in Section~\ref{sec:evidence} is what does.}
  \label{fig:method-comparison}
\end{figure}

\textbf{Comparison with a span-supervised system.} Span NLI BERT \citep{koreeda2021contractnli}, trained with the full evidence layer (all 559,215 span decisions), reports 0.892 accuracy on this split; label-only SFT reaches 0.896 on the recorded verdicts alone, and RSFT 0.797. We flag this only as a descriptive, unmatched comparison --- different architecture, different supervision (span-level vs verdict-only), and an evaluation convention we did not reproduce --- and claim no equivalence to a span-supervised system. The point is narrow: a 9B model post-trained on bare verdicts reaches verdict accuracy comparable to a fully span-supervised system here, which is also what most sharpens the contamination question (\hyperref[sec:limitations]{Limitations}).

\subsection{Error analysis}
\label{sec:error}

\begin{figure*}[t]
  \centering
  \includegraphics[width=\linewidth]{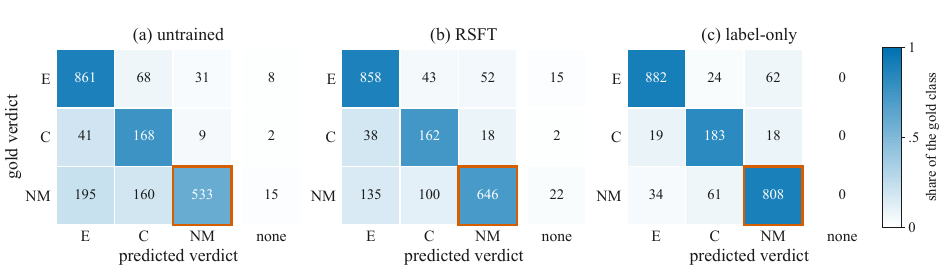}
  \caption{Verdict confusion (matched decode). The fourth predicted column, \texttt{none}, holds outputs the parser could not resolve to a verdict (25 / 39 / 0); these score as incorrect, and including them is what makes every panel sum to all 2,091 test rows. The NotMentioned diagonal grows from the untrained student (533) to RSFT (646) to label-only (808); label-only also improves the Entailment and Contradiction diagonals, whereas RSFT trades a little E/C recall for its NM recovery.}
  \label{fig:confusion}
\end{figure*}

Total errors under RSFT fall from 529 to 425. Decomposing the $-$104-row net change by gold class states what the aggregate hides:

\begin{table}[t]
  \centering
  \footnotesize
  \setlength{\tabcolsep}{4pt}
  \begin{tabular}{@{}p{0.44\columnwidth}p{0.12\columnwidth}p{0.24\columnwidth}@{}}
    \toprule
    \textbf{Row population} & \textbf{n} & \textbf{Change in correct predictions} \\
    \midrule
    Gold NotMentioned & 903 & +113 \\
    Gold Entailment or Contradiction (the rows the evidence metric scores) & 1,188 & $-$9 \\
    Net & 2,091 & +104 \\
    \bottomrule
  \end{tabular}
  \caption{Where RSFT's accuracy gain comes from.}
  \label{tab:gain}
\end{table}

The entire net gain is NotMentioned recovery; on the 1,188 evidence-bearing rows RSFT is 9 predictions \emph{worse}. Predicted NotMentioned moves 27.4\% $\rightarrow$ 34.2\% against a gold rate of 43.2\% --- a shift of one class marginal toward the corpus rate. Entailment recall falls 0.889 $\rightarrow$ 0.886 and Contradiction recall 0.764 $\rightarrow$ 0.736; the per-class F1 gains for E and C are precision gains from fewer spurious E/C predictions on NotMentioned rows, i.e. the same phenomenon as the NM recovery, not three independent improvements. Label-only SFT, by contrast, raises NotMentioned recall much further (to 0.895) and lands its predicted NM share on the gold marginal (0.425 vs 0.432) while E and C recall \emph{rise} rather than fall (E 0.889 $\rightarrow$ 0.911, C 0.764 $\rightarrow$ 0.832) --- which is why its accuracy lands higher without the E/C regression RSFT shows.

\textbf{Two confounds favour the label-only arm.} Both are inherent to the cheaper method rather than measurement artifacts, but the headline +0.099 accuracy gap should be read with them in mind. (i) The arms do not train on the same rows. RSFT trains on 6,251 rows drawn from the 6,938 Gate-1-covered ones; label-only trains on 6,477 from the full 7,191, so it sees --- with gold labels --- the 253 rows RSFT structurally cannot cover, which are the hardest rows (no sampled trace ever produced the recorded verdict), concentrated on nda-2/16/17. This shows in the data: on nda-2/16/17 label-only leads RSFT by +0.21 (0.870 vs 0.661), against +0.08 on the other 14 hypotheses (0.901 vs 0.826), so coverage is doing part of the work. (ii) Label-only cannot truncate. Its 71-character targets never reach the 900-token cap, so it has zero unparsed rows against 25 (untrained) and 39 (RSFT), and unparsed scores as wrong; restricted to the 2,052 rows RSFT parses, RSFT is 0.812 and label-only 0.897, a gap of +0.085 --- so roughly 1.4 of the 9.9 points is parse-failure asymmetry rather than verdict quality.

\subsection{How much of the accuracy gain is calibration?}
\label{sec:calibration}

If the accuracy gain is mostly a class-marginal correction, a label re-mapping that reads no contract should reproduce it. We fit such a control on the untrained student's own training-split predictions and apply it to the test predictions; results are intersected on the rows both systems parse (2,025 rows), since a relabelling control has no label to re-map on an unparsed row.

\begin{table}[t]
  \centering
  \footnotesize
  \setlength{\tabcolsep}{4pt}
  \begin{tabular}{@{}p{0.34\columnwidth}p{0.16\columnwidth}p{0.16\columnwidth}p{0.18\columnwidth}@{}}
    \toprule
    \textbf{System} & \textbf{Accuracy} & \textbf{vs untrained} & \textbf{Share of the RSFT gain} \\
    \midrule
    untrained student & 0.764 & --- & --- \\
    + global calibration & 0.764 & +0.000 & 0\% \\
    + per-hypothesis calibration & 0.782 & +0.018 & 37\% \\
    lookup control (no model) & 0.667 & $-$0.098 & --- \\
    RSFT student & 0.813 & +0.048 & 100\% \\
    \bottomrule
  \end{tabular}
  \caption{Calibration control (intersected population, unparsed dropped).}
  \label{tab:calibration}
\end{table}

A label re-mapping recovers about a third of the increment, not most of it. And it cannot reproduce the \emph{manner} of the shift: calibration buys NotMentioned recall by spending precision (NM precision/recall 0.928/0.613 $\rightarrow$ 0.825/0.692), whereas RSFT gains recall and keeps precision (0.904/0.732). The remainder of the increment is not explained by the label marginal. (The absolute delta here, +0.048, differs from the +0.050 in Table~\ref{tab:paired} only because this population drops unparsed rows; the \emph{share} of the gain is the stable quantity to quote. For the same reason the lookup control reads 0.667 in this table, against 0.659 on the full test split reported elsewhere --- the two figures are the same control on the intersected versus the full population.)

\section{Evidence quality without evidence annotation}
\label{sec:evidence}

The central evidence result is worth stating sharply. A model fine-tuned only on bare verdicts, under a prompt that contains no evidence paragraph, and then evaluated under the citation-requesting \texttt{faithful} prompt, produces \emph{higher} span agreement with withheld human evidence than the untrained model --- 0.564 vs 0.493. Evidence quality it was never trained to produce improves. This is the cleanest support for the verdict-mediation reading below, and it holds without any evidence annotation entering training.

Evidence is evaluated on the 1,188 rows with human-marked spans (968 entailment, 220 contradiction). We lead with source verifiability --- the share of rows whose every quotation is found verbatim in the contract --- because it is the one metric independent of the grounding instrument used in selection (Section~\ref{sec:rsft}). Span F1 against withheld human spans is reported second.

\begin{table}[t]
  \centering
  \footnotesize
  \setlength{\tabcolsep}{4pt}
  \begin{tabular}{@{}p{0.31\columnwidth}p{0.18\columnwidth}p{0.13\columnwidth}p{0.20\columnwidth}@{}}
    \toprule
    \textbf{Measure} & \textbf{Untrained} & \textbf{RSFT} & \textbf{Label-only} \\
    \midrule
    Rows all-verbatim (source verifiability) & 59.7\% & 70.1\% & 72.9\% \\
    Verbatim rate among rows that cite & 61.8\% & 75.8\% & 78.0\% \\
    Span precision vs human gold & 0.830 & 0.859 & 0.882 \\
    Span recall vs human gold & 0.351 & 0.411 & 0.415 \\
    Span F1 vs human gold & 0.493 & 0.556 & 0.564 \\
    Rows emitting any citation & 96.5\% & 92.5\% & 93.4\% \\
    \bottomrule
  \end{tabular}
  \caption{Evidence quality against withheld human annotation, matched decode, \texttt{faithful} prompt.}
  \label{tab:evidence}
\end{table}

\textbf{Both methods recover evidence quality without any evidence annotation.} As a gain over the untrained faithful baseline, span F1 rises +0.062 for RSFT (95\% paired bootstrap CI [+0.042, +0.082]) and +0.071 for label-only ([+0.053, +0.090]); both exclude zero. On the evidence axis the \emph{two fine-tunes} are statistically indistinguishable: the label-only-minus-RSFT span-F1 difference is +0.009 with a 95\% CI of [$-$0.011, +0.028] (document-clustered [$-$0.014, +0.031]), spanning zero, and they differ by only 2.8 points of all-verbatim rate. The result we lean on is therefore not a gap between the two arms --- there is none we can resolve at one seed --- but that a model which never saw a citation in training recovers evidence quality at all.

\textbf{Why, mechanically --- and the two arms differ here.} Evidence F1 is mediated by the verdict: a NotMentioned prediction returns empty spans, a correct Entailment/Contradiction prediction returns citations, and the base model already extracts evidence well (untrained 0.49 F1, 96.5\% of rows cite). Label-only's gain fits this cleanly --- it gets more E/C verdicts right (Section~\ref{sec:error}), so it emits citations on more of the scored rows. RSFT's does not: on the 1,188 E/C rows its verdicts are 9 \emph{worse} and it cites on fewer of them (92.5\% vs the base's 96.5\%), yet it recovers 143 more gold spans (true positives 844 $\rightarrow$ 987). Its span-F1 gain therefore comes from better citations on the rows it does cite --- from the source-grounded rationale supervision --- not from the verdict. The two arms reach almost the same span F1 (0.556 vs 0.564) by different routes: label-only through the verdict, RSFT through citation quality. So verdict mediation carries the label-only result but not the RSFT one; neither route requires evidence annotation.

\begin{table}[t]
  \centering
  \footnotesize
  \setlength{\tabcolsep}{4pt}
  \begin{tabular}{@{}p{0.23\columnwidth}p{0.07\columnwidth}p{0.18\columnwidth}p{0.11\columnwidth}p{0.20\columnwidth}@{}}
    \toprule
    \textbf{Class} & \textbf{n} & \textbf{Untrained} & \textbf{RSFT} & \textbf{Label-only} \\
    \midrule
    Entailment & 968 & 0.506 & 0.564 & 0.572 \\
    Contradiction & 220 & 0.436 & 0.517 & 0.530 \\
    \bottomrule
  \end{tabular}
  \caption{Evidence F1 by class (\texttt{faithful} prompt).}
  \label{tab:evidence-class}
\end{table}

\textbf{Locatable subset.} Restricting to the 1,037 rows where all three systems emit a string-locatable citation --- which removes the locatability component the shared instrument otherwise credits --- the ranking holds: untrained 0.510, RSFT 0.586, label-only 0.590. This subset conditions on a post-treatment variable --- all three systems emitting a locatable citation --- so it excludes exactly the rows where RSFT flipped its verdict to NotMentioned and cited nothing; the comparison is descriptive of the shared-citation rows, not an unbiased treatment effect. The evidence advantage is not solely a locatability effect.

\textbf{Span-matching and scoring.} A citation is matched to gold spans by the same three-tier ladder used in selection (Section~\ref{sec:rsft}): exact string, whitespace/punctuation-normalised retry, then credit if $\geq$60\% of its content tokens fall in one contiguous stretch. Precision, recall, and F1 are pooled counts over the 1,188 rows, not per-row averages. A citation that cannot be located anywhere in the contract contributes one false positive and no true positive, so it lowers precision; an eligible row that emits no citation lowers recall. Unlocatable citations are rare (3--4 excerpts per system). Pooled counts are tp/fp/fn = 844/173/1560 (untrained), 987/162/1417 (RSFT), 998/134/1406 (label-only). Mean citations per citing row (1.11--1.19) and mean citation length (223--233 characters) are similar across arms, so the recall differences are not driven by longer or more numerous quotes.

\textbf{What the comparison supports, and what it does not.} We do not claim rejection sampling is the reason evidence quality improves: a cheaper intervention --- supervised fine-tuning on the bare labels the record already contains --- reaches comparable evidence quality in this run, so the rejection-sampling machinery is not established as necessary for the gain. Nor do we claim label-only SFT is the better method; a single seed cannot support that ordering. What the comparison does support is the weaker and, we think, more useful statement: evidence quality is recoverable from verdict-only records without evidence annotation, and both methods achieve this using supervision derived from the recorded verdict. It is worth noting which arm had the richer inputs: RSFT drew about a fifth of its selected traces from a 31B sampler (Table~\ref{tab:sources}), while label-only used nothing beyond the recorded labels, and the two still land within a non-significant 0.009 of each other on span F1.

\section{Verdict accuracy and evidence quality decouple}
\label{sec:decouple}

\begin{figure}[t]
  \centering
  \includegraphics[width=\columnwidth]{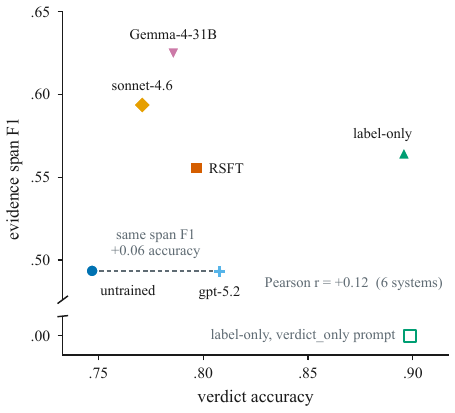}
  \caption{Cross-system relationship between verdict accuracy and evidence span F1, six systems under the \texttt{faithful} prompt. The two axes are only weakly related (descriptive Pearson +0.12 for span F1, +0.20 for verbatim rate) and induce different rankings: gpt-5.2 has the highest off-the-shelf accuracy yet is no better on evidence than the untrained student (span F1 0.493 for both --- the pair the on-figure annotation marks), while the 31B sampler and sonnet-4.6 are lower-accuracy yet highest on evidence (the label-only fine-tune is the most accurate system overall, at 0.896). The open square is label-only under the \texttt{verdict\_only} prompt (accuracy 0.899, span F1 0.000 by construction --- no citation is requested), the extreme point of the decoupling; it is excluded from the correlation, and the vertical axis is broken below 0.50 so that it can be shown. The 31B sampler was decoded at a 700-token cap (all other systems at 900), but only 1 of its 2,091 rows reached the cap, so its position is effectively unaffected; see Table~\ref{tab:decoupling}.}
  \label{fig:decoupling}
\end{figure}

\begin{table*}[t]
  \centering
  \small
  \begin{tabular}{@{}lrrr@{}}
    \toprule
    \textbf{System} & \textbf{Accuracy} & \textbf{Span F1} & \textbf{Verbatim rate} \\
    \midrule
    Qwen3.5-9B, untrained & 0.747 & 0.493 & 59.7\% \\
    sonnet-4.6 & 0.771 & 0.594 & 70.9\% \\
    Gemma-4-31B (sampler) & 0.786 & 0.625 & 77.5\% \\
    Qwen3.5-9B, RSFT & 0.797 & 0.556 & 70.1\% \\
    gpt-5.2 & 0.808 & 0.493 & 48.7\% \\
    Qwen3.5-9B, label-only & 0.896 & 0.564 & 72.9\% \\
    \bottomrule
  \end{tabular}
  \caption{\textbf{Verdict accuracy and evidence quality across the six systems} (\texttt{faithful} prompt), ordered by accuracy. The two axes rank systems differently: accuracy rises down the column while span F1 and verbatim rate do not track it. All open-weight/student rows use the 900-token cap except the 31B sampler, which was decoded at 700; only 1 of its 2,091 rows reached that cap, so its figures are effectively cap-insensitive (the untrained student, with 62 rows at the 700 cap, is the one re-run at 900; Section~\ref{sec:config}). The two commercial systems do not record completion-token counts, so their caps cannot be read from the saved records.}
  \label{tab:decoupling}
\end{table*}

The rank reversals are visible directly in the measurements and do not depend on the correlation coefficients, which with six systems summarize this set rather than a population; the decoupling rests on the rankings, not the exact coefficient. The coefficients are moreover composition-dependent: across the four off-the-shelf systems alone the accuracy--verbatim correlation is $-$0.27, and adding the two fine-tunes moves it to +0.20 --- a further reason to read the rankings, not the number. gpt-5.2 is the plainest counterexample --- highest off-the-shelf accuracy, lowest verbatim rate, and span F1 tied with the untrained student at the bottom of the column --- and its cap is not the cause, since its accuracy, not its evidence, is what leads. Label-only SFT under the \texttt{verdict\_only} prompt is the cleanest illustration of the decoupling: accuracy 0.899 with span F1 0.000 by construction --- high verdict accuracy and no evidence at all, from one system, because the prompt requests no citation (the open square in Figure~\ref{fig:decoupling}).

The practical reading: for workflows that need reviewable evidence, verdict accuracy alone is insufficient for model selection. A system can produce more correct labels while providing quotations that are less often verifiable or less aligned with withheld human evidence. And within a single model, whether it emits evidence at all is a property of the prompt, not only of the weights (\hyperref[sec:limitations]{Limitations}).

\section{Secondary analyses}
\label{sec:secondary}

\textbf{Sampling recovery at matched budget.} Among rows both models miss under greedy decoding, the 9B student recovers 52.0\% at the matched sampling budget vs 13.2\% for the 31B sampler (n = 994, matched k = 8, unbiased estimator $1 - C(n-c,k)/C(n,k)$). These are point estimates without a confidence interval, computed over the training-split candidate generation rather than the test evaluation; they show standalone greedy accuracy does not determine recovery under sampling, but do not establish why the rates differ and are not a claim about diversity.

\textbf{Training-source mixture (RSFT).} The sampler's share of selected traces varies across hypotheses from 2\% to 46\%. The descriptive correlation between the student's training-trace share and its test-accuracy gain is $-$0.14 across 17 hypotheses; hypotheses below the median student share gain 0.069 on average vs 0.033 at or above. With 17 hypotheses this supports no directional inference and isolates no component; a matched ablation is required.

\textbf{Hypothesis-only control.} A control reading hypothesis identity but no contract reaches 0.659, consistent with hypothesis-specific label priors explaining a substantial share of performance. Performance above it is consistent with using more than those priors but does not exclude memorisation or pretraining exposure.

\section{Experimental configuration}
\label{sec:config}

Both fine-tunes use the configuration in Table~\ref{tab:training}; label-only SFT differs only in the target text and the training prompt. The recipe was fixed before training; no hyperparameter was tuned on the test split and no checkpoint was selected on test results.

\begin{table*}[t]
  \centering
  \small
  \begin{tabular}{@{}p{0.22\textwidth}p{0.72\textwidth}@{}}
    \toprule
    \textbf{Setting} & \textbf{Value} \\
    \midrule
    Base model & Qwen/Qwen3.5-9B, Apache 2.0, 4-bit (Unsloth; \citealp{unsloth}) \\
    Adapter & QLoRA, rank 64, $\alpha$ 128, dropout 0.05, no bias \\
    Target modules & q\_proj, k\_proj, v\_proj, o\_proj, gate\_proj, up\_proj, down\_proj \\
    Loss & Cross-entropy on target tokens only; prompt labels set to $-$100 manually \\
    Learning rate & $5\times 10^{-5}$, cosine decay, 20 warmup steps \\
    Weight decay & 0.05 \\
    Batch & 1 $\times$ grad-accum 8 = effective batch 8 \\
    Epochs / steps & 1 epoch; RSFT 782 steps (n\_train = 6,251), label-only 810 steps (n\_train = 6,477) \\
    Max sequence length & 16,384 tokens, keep-end truncation \\
    Seed & 42, single seed per arm \\
    Hardware & 1 $\times$ L40S, 48 GB \\
    Training wall-clock & RSFT 2.90 h, 13.4 s/step; label-only 2.63 h, 11.7 s/step. Label-only runs more
steps on more rows yet finishes sooner: its 71-character targets cost about 12\%
less per step than RSFT's 1,294-character traces, against the shared
\textasciitilde{}13.6k-token prompt \\
    Checkpoint selection & None; final step retained \\
    Evaluation decoding & vLLM \citep{10.1145/3600006.3613165}, greedy, single pass, 900-token cap for every run except the 31B sampler (700; see Section~\ref{sec:eval}) \\
    \bottomrule
  \end{tabular}
  \caption{Training configuration. Hyperparameters are shared by both methods; training rows, step count and wall-clock differ by arm.}
  \label{tab:training}
\end{table*}

\textbf{Robustness: a second epoch does not close the gap.} A 2-epoch RSFT run (\texttt{cnli-rsft-qwen-e2}) reaches accuracy 0.796 and span F1 0.555 --- indistinguishable from the 1-epoch run (0.797 / 0.556). This argues against insufficient epoch count as the explanation for the gap to label-only SFT; it does not rule out other differences in the training schedule. A second training seed per arm remains outstanding and single-seed point estimates should be read as provisional (\hyperref[sec:limitations]{Limitations}, Section~\ref{sec:roadmap}).

\textbf{Pre-registration (legacy).} Our code carries a pre-registration from an earlier feasibility framing, with targets fixed before training (accuracy $\geq$ 0.850, all-verbatim rate $\geq$ 0.700, span F1 $\geq$ 0.600); of these the RSFT recipe met only the verbatim target (0.701). We report this for completeness, but these absolute deployment gates are not the object of this study, whose claims concern the relationship between the two quality axes rather than clearing a fixed threshold.

\textbf{Statistics and multiplicity.} Confidence intervals are 2.5/97.5 percentile paired bootstraps (10,000 resamples; seed 37 for accuracy, seed 6 for span F1) \citep{tibshirani1993introduction}, reported row-level and document-clustered; verdict comparisons also use the exact two-sided McNemar test \citep{McNemar_1947}. The paper runs several paired tests without a multiplicity correction: the headline McNemar p-values ($\leq 4\times 10^{-12}$) survive any reasonable correction, but the per-class F1 movements (Section~\ref{sec:error}), the cross-system correlation (Section~\ref{sec:decouple}), and the sampling-recovery figures (Section~\ref{sec:secondary}) are uncorrected point estimates and should be read as descriptive.

\textbf{Matched-decode correction.} An earlier release stated that decoding was identical for every system. Read off the completion-token counts, it was not: the greedy family (untrained student, 31B sampler) was capped at 700 while the post-trained student, the sampled passes, and the frontier runs used 900. We re-ran the untrained student at 900 (\texttt{cnli-greedy-qwen-m900}); the 31B sampler was not re-run, and Section~\ref{sec:eval} and Table~\ref{tab:decoupling} state why its figures are cap-insensitive. The accuracy delta is unchanged to four decimals (+0.0497) and the evidence delta is slightly \emph{larger} under the matched cap (+0.0586 $\rightarrow$ +0.0622); only the baseline moved, because only the baseline was decoded unfairly. We record the completion-token count and the token cap on every generated record, alongside the prompt hash \texttt{f9318d34}.

\textbf{Data (RSFT).} 6,251 training and 687 validation traces, one per row, split by document (381 train, 42 validation documents); target length median 312 tokens. Assembled class mix 50.2 / 11.8 / 38.0 vs corpus 49.1 / 11.7 / 39.2 --- every class within 1.2 points. Gate-1 coverage by class: 98.6\% entailment, 97.6\% contradiction, 93.4\% not-mentioned; the 253 uncovered rows concentrate on nda-2/16/17 (153 of them). Validation loss is used for monitoring only and is flat over the final interval (0.1822 $\rightarrow$ 0.1823), not monotone to the end.

\textbf{Data (label-only).} 6,477 training and 714 validation rows on the same
381 / 42 document split, with every ContractNLI row retained: this arm needs no
candidate to survive Gate~1. Targets average 71 characters, and training ran 810
steps in 2.63 h.

\section{Relation to prior work}
\label{sec:related}

\textbf{Rejection-sampling and self-training fine-tuning.} The method family that RSFT belongs to bootstraps supervision from a model's own filtered generations: STaR \citep{NEURIPS2022_639a9a17} iterates rationale generation, correctness filtering, and retraining; RFT \citep{yuan2023scalingrelationshiplearningmathematical} filters sampled reasoning paths on answer correctness and RAFT \citep{dong2023raft} reward-ranks sampled generations under a learned reward model; Llama 2 \citep{touvron2023llama2openfoundation} uses rejection sampling in its alignment pipeline; ReST \citep{gulcehre2023reinforcedselftrainingrestlanguage} frames the loop as reinforced self-training; and Distilling Step-by-Step \citep{hsieh2023distilling} supervises rationales from a larger teacher. Our RSFT is one round of verdict-filtered, source-grounded rejection sampling, and label-only SFT is the direct-answer baseline against which these methods are usually measured \citep{NEURIPS2022_639a9a17,hsieh2023distilling}. Our two gates correspond to these two operations --- Gate 1 filters on the recorded verdict, Gate 2 ranks the survivors by an automatic grounding score. The observation that the direct-answer baseline reaches comparable evidence quality, at higher verdict accuracy, is specific to this recipe on this task and not a claim about the family; Section~\ref{sec:roadmap} lays out combinations we have not tried: stacking, iteration, and preference finishing \citep{NEURIPS2023_a85b405e}. We build on LoRA \citep{hu2022lora} and QLoRA \citep{NEURIPS2023_1feb8787} for adaptation and serve with vLLM \citep{10.1145/3600006.3613165}.

\textbf{NLI artifacts.} The hypothesis-only lookup control re-derives the class-prior artifact that hypothesis-only baselines expose in NLI datasets \citep{poliak2018hypothesis,gururangan2018artifacts}; reporting it is important when the same requirement set repeats across documents.

\textbf{Attribution and citation quality.} Evaluating whether model citations are verifiable and support the claim connects to work on generation with citations \citep{gao2023alce}, researching-and-revising attributions \citep{gao2023rarr}, and verifiability in generative search \citep{liu2023verifiability}. Those evaluate model- or NLI-judged support of free-text claims; we instead measure agreement with \emph{withheld human evidence spans} and verbatim source verifiability, which suits the verdict-only, span-annotated setting of ContractNLI.

Bridgewater AIA Labs and Thinking Machines \citep{su2026expertjudgment} report that expert judgment on financial triage can be learned by an open-weight model and deployed; their pipeline commissions labels through paid annotation and trains Qwen3-235B with reinforcement learning (GRPO, then a CISPO loss with asymmetric clipping) and on-policy distillation. We differ in that the verdict labels already exist rather than being commissioned, the human evidence layer is withheld rather than purchased, and the recipe is single-accelerator SFT rather than RL. The comparison motivates the focus on recorded decisions but is not a matched baseline.

\section{Conclusion}
\label{sec:conclusion}

In verdict-only records, evidence quality is recoverable by post-training with no human evidence annotation. The two axes decouple across systems, yet both improve under verdict-only fine-tuning. We compared two verdict-only methods --- a bare label-only SFT and an elaborate verdict-filtered, source-grounded rejection-sampling recipe --- and both recover evidence quality; in this single run they land close on the evidence axis and further apart on accuracy, and a 2-epoch RSFT run reproduces the 1-epoch result, arguing against insufficient epoch count as the explanation for the gap. We do not rank the two methods: the estimates are single-seed on one corpus and one student, and the comparison is a starting point, not a verdict. What it does establish is that neither method needs evidence annotation, and that the same evidence quality is reached by two routes --- label-only through the verdict, RSFT through the better citations its source-grounded supervision buys (Section~\ref{sec:evidence}).

\section{Roadmap and future work}
\label{sec:roadmap}

This work establishes the two-axis framing and a clean comparison; the following extensions turn it into a program. Each is feasible on the same single-accelerator budget.

\begin{enumerate}
\item \textbf{Second seeds and a wider model family.} Run each arm at three seeds and report mean $\pm$ s.d.; repeat on a second student (a different 7--9B base) to separate method effects from a single checkpoint's idiosyncrasies. Required before any point estimate here is stated as a headline.
\item \textbf{A minimal-prompt citation test.} Evaluate every fine-tune under \texttt{verdict\_only} \emph{and} \texttt{faithful} to measure how much citation behaviour is weight-resident vs prompt-elicited --- the one dimension on which RSFT is expected to hold an advantage over label-only.
\item \textbf{Preference finishing (DPO/ORPO).} Use the grounding score and verdict-correctness to construct preference pairs and add a preference-optimisation stage \citep{NEURIPS2023_a85b405e} after SFT, targeting the E/C recall that both current fine-tunes leave on the table.
\item \textbf{Semantic evidence credit.} Add an embedding- or entailment-based span-matching rule beside the string ladder, so evidence quality is not measured with the same instrument used in selection.
\item \textbf{More corpora.} Add one or two additional verdict-plus-withheld-evidence datasets (e.g. other document-level NLI or claim-verification corpora) to test whether the verdict-mediation mechanism generalizes beyond ContractNLI.
\item \textbf{Verifier and lookup ablations.} Replace the recorded-verdict filter with the sampler pool's modal answer to estimate the contribution of the recorded verdict, and report the companion lookup constructions.
\end{enumerate}

\section*{Limitations}
\label{sec:limitations}

\textbf{One corpus, one student, one seed per arm.} Intervals quantify variation over test rows, not over training initialisations. A second seed per arm is outstanding; the +0.099 label-only-vs-RSFT accuracy gap is large relative to plausible seed variance but is a single-seed estimate.

\textbf{Label-only's citations are prompt-elicited, not weight-resident.} Label-only SFT emits evidence only because the \texttt{faithful} prompt asks for it; under \texttt{verdict\_only} it cites nothing. RSFT trains citation behaviour into the target, so it should cite under a minimal prompt where label-only would not --- a difference we have not tested (Section~\ref{sec:roadmap}). In this study's protocol (the \texttt{faithful} prompt) label-only cites at 93.4\%, so the practical difference is nil here.

\textbf{Evidence F1 is partly a locatability measure.} Selection and evaluation share the same three-tier string-matching ladder; the verbatim rate is the non-circular metric and we lead with it. The locatable-subset result (Section~\ref{sec:evidence}) shows the ranking survives removing the shared instrument, but a semantic credit rule (embedding- or entailment-based) is future work.

\textbf{Pretraining exposure cannot be excluded.} ContractNLI has been public since 2021 and its evidence layer (559,215 binary span decisions) is public; fine-tuning can surface memorised content. This is most pressing for the label-only arm's 0.896 accuracy --- the result closest to the supervised figure (Section~\ref{sec:aggregate}) --- which is the one that most needs the closed-book probe, with and without document identity, that we leave to future work. Paired comparisons control for fixed differences between checkpoints on the same rows but do not eliminate contamination.

\textbf{Component effects are not isolated within RSFT.} Verdict filtering, grounding selection, self-training, and sampler contribution are evaluated as one pipeline.

\bibliography{references}

\appendix

\section{Prompts and generation}
\label{sec:appendix}

\subsection{The two prompts}
\label{sec:prompts}

Everything above \texttt{\#\# OUTPUT} is byte-identical between the two variants except the single evidence paragraph. Hashes: \texttt{faithful f9318d34}, \texttt{verdict\_only 7ada9e8d}.

\textbf{\texttt{faithful}} --- full prompt (reasoning + citations); used by the main system and every evidence-axis evaluation:

\begin{lstlisting}[basicstyle=\ttfamily\footnotesize]
You are given a contract and a hypothesis.

## CONTRACT
{document}

## HYPOTHESIS
{hypothesis}

## TASK
Classify whether the hypothesis is entailed by, contradicting to, or not
mentioned by (neutral to) the contract. Answer with exactly one of:
Entailment, Contradiction, NotMentioned.

Identify the evidence for your decision as quotes from the contract. A piece
of evidence is a sentence, or a list item within a sentence. Evidence should
be as concise as possible while being self-contained, so that a reader can
understand it from the quote alone. Quote it verbatim. Evidence is only
defined when the answer is Entailment or Contradiction; for NotMentioned,
return an empty list.

## OUTPUT
Reason step by step, then return your final answer as a single block:
<answer>
{{"hypothesis_id": "{hypothesis_id}", "answer": "Entailment | Contradiction | NotMentioned", "reasoning": "2-3 sentence explanation", "confidence_score": <1-10>, "source_attribution": [{{"excerpt": "verbatim quote from the contract"}}]}}
</answer>
The <answer> block must contain valid JSON only.
\end{lstlisting}

\textbf{\texttt{verdict\_only}} --- \texttt{faithful} with the evidence paragraph removed and the output reduced to the verdict; used to train and evaluate label-only SFT on the accuracy axis:

\begin{lstlisting}[basicstyle=\ttfamily\footnotesize]
You are given a contract and a hypothesis.

## CONTRACT
{document}

## HYPOTHESIS
{hypothesis}

## TASK
Classify whether the hypothesis is entailed by, contradicting to, or not
mentioned by (neutral to) the contract. Answer with exactly one of:
Entailment, Contradiction, NotMentioned.

## OUTPUT
Return your answer as a single block:
<answer>
{{"hypothesis_id": "{hypothesis_id}", "answer": "Entailment | Contradiction | NotMentioned"}}
</answer>
The <answer> block must contain valid JSON only.
\end{lstlisting}

\subsection{Candidate generation (RSFT)}
\label{sec:generation}

Candidates are produced in two passes. Greedy first from both models on every training row; then, only on rows where greedy produced nothing the verdict filter would accept, a sampled pass at temperature 1.0. Cost scales with the residue rather than the corpus.

\textbf{Decoding settings.} The greedy pass is $k = 1$, $T = 0$ for both models. The sampled pass runs only on rows whose greedy output the verdict filter rejected:

\begin{itemize}
\itemsep0em
  \item \textbf{Student (Qwen3.5-9B):} $k = 32$, $T = 1.0$, top-$p$ 0.95, top-$k$ 20.
  \item \textbf{Sampler (Gemma-4-31B):} $k = 8$, $T = 1.0$, top-$p$ 0.95, top-$k$ 64.
\end{itemize}

Both models use a 900-token completion cap.

Rows are ordered document-major so all 17 hypotheses about one contract are answered while it is in the prefix cache, which is most of why the whole generation fits in 6.8 GPU-hours. Evaluation decoding is greedy, single pass, 900-token cap for every system.

\end{document}